%% file: ijcai26.tex
\documentclass{article}
\usepackage{ijcai26}

\usepackage{times}
\usepackage{soul}
\usepackage{url}
\usepackage[hidelinks]{hyperref}
\usepackage[utf8]{inputenc}
\usepackage[small]{caption}
\usepackage{graphicx}
\usepackage{amsmath}
\usepackage{amsthm}
\usepackage{booktabs}
\usepackage{algorithm}
\usepackage{algorithmic}
\usepackage[switch]{lineno}

\usepackage{amssymb,amsfonts}
\usepackage{adjustbox}
\usepackage{multirow}
\usepackage{array}
\usepackage{pifont}
\usepackage{tikz}
\usetikzlibrary{shapes.geometric}
\usepackage{siunitx}
\usepackage[table]{xcolor}
\usepackage{colortbl}
\usepackage{subcaption}
\usepackage{float}

\newcommand{\greencheck}{\textcolor{green!60!black}{\ding{51}}}
\newcommand{\redx}{\textcolor{red!70!black}{\ding{55}}}
\definecolor{oursrow}{RGB}{235,245,255} 

\newcolumntype{C}[1]{>{\centering\arraybackslash}m{#1}}

\newcommand{\symline}{0.6pt}
\newcommand{\symsize}{1.2ex}
\newcommand{\symbase}{0.0ex}

\newcommand{\qtensor}{%
  \tikz[baseline=\symbase]\draw[line width=\symline]
    (0,0.866*\symsize) -- (\symsize,0.866*\symsize) -- (0.5*\symsize,0) -- cycle;}

\newcommand{\qtoken}{%
  \tikz[baseline=\symbase]\draw[line width=\symline]
    (0,0) -- (\symsize,0) -- (0.5*\symsize,0.866*\symsize) -- cycle;}

\newcommand{\qhead}{%
  \tikz[baseline=\symbase]\draw[line width=\symline]
    (0.5*\symsize,0.5*\symsize) circle [radius=0.48*\symsize];}

\newcommand*\circnum[1]{%
  \tikz[baseline=(char.base)]
    \node[draw,circle,inner sep=0.2pt,minimum size=1.8ex,font=\scriptsize] (char) {#1};}

\title{QFlash: Bridging Quantization and Memory Efficiency in Vision Transformer Attention}

\author{
    Sehyeon Oh$^{1,2}$
    \and
    Yongin Kwon$^3$
    \and
    Jemin Lee$^4$\footnote{Corresponding author}
    \affiliations
    $^1$University of Science and Technology, Daejeon, Republic of Korea\\
    $^2$Electronics and Telecommunications Research Institute, Daejeon, Republic of Korea\\
    $^3$Pusan National University, Busan, Republic of Korea\\
    $^4$Jeonbuk National University, Jeonju-si, Republic of Korea
    \emails
    somae0604@gmail.com,
    yongin@pusan.ac.kr,
    jemin.lee@jbnu.ac.kr
}

\begin{document}

\maketitle

\begin{abstract}

FlashAttention improves efficiency through tiling, but its online softmax still relies on floating-point arithmetic for numerical stability, making full quantization difficult.
We identify three main obstacles to integer-only FlashAttention: 
(1) scale explosion during tile-wise accumulation, 
(2) inefficient shift-based exponential operations on GPUs, and 
(3) quantization granularity constraints requiring uniform scales for integer comparison.
To address these challenges, we propose \textit{QFlash}, an end-to-end integer FlashAttention design that performs softmax entirely in the integer domain and runs as a single Triton kernel.
On seven attention workloads from ViT, DeiT, and Swin models, QFlash achieves up to 6.73$\times$ speedup over I-ViT and up to 8.69$\times$ speedup on Swin, while reducing energy consumption by 18.8\% compared to FP16 FlashAttention, without sacrificing Top-1 accuracy on ViT/DeiT and remaining competitive on Swin under per-tensor quantization. 
Our code is publicly available at \url{https://github.com/EfficientCompLab/qflash}.
\end{abstract}

\section{Introduction}

Transformer self-attention provides strong representational power, but its computation and memory cost grow as $O(N^2)$ with sequence length.
This growth incurs a memory bottleneck in Vision Transformers (ViTs), where intermediate tensors must be transferred between on-chip and off-chip memory.
FlashAttention~\cite{dao2022flashattention} was proposed to reduce this problem by splitting the sequence into tiles.
Each tile is computed on-chip, and only the final outputs are written to off-chip memory.

However, FlashAttention and later works~\cite{dao2022flashattention,dao2023flashattention,shah2024flashattention} cannot compute the softmax in one step because of the tile-based design.
They update the row-wise maximum for each tile, rescale the past results, and then add the new tile results.
This process needs numerical stability and therefore depends on floating-point operations.

Other works on quantizing attention include I-BERT~\cite{kim2021bert}, I-ViT~\cite{li2023vit}, QAttn~\cite{kluska2024qattn}, and INT-FlashAttention~\cite{chen2024int}.
Although I-BERT and I-ViT quantize the Softmax operation, they do not consider a tile-based fused attention design like FlashAttention~\cite{dao2022flashattention}, leaving memory bottlenecks unresolved.
QAttn and INT-FlashAttention quantize only the matrix multiplications, while the softmax still runs in floating point.
As a result, there has been no work on a fully integer-only fused attention that reduces memory cost and removes floating-point operations.

\begin{figure}[t]
\centering
\includegraphics[width=0.48\textwidth]{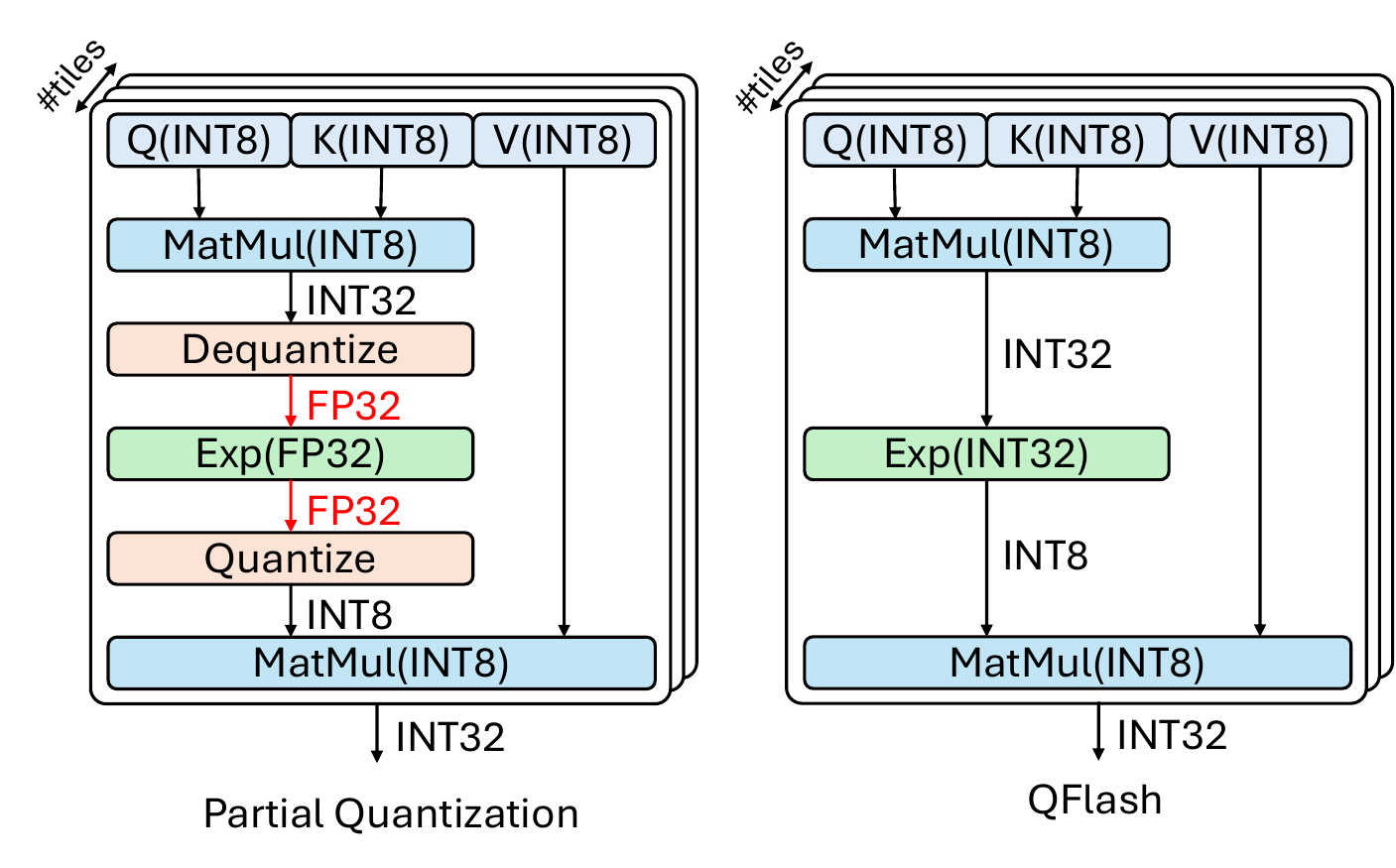}
\caption{Comparison of partial quantization and the proposed QFlash method.}
\label{fig:1}
\end{figure}

Through our analysis of integer-only FlashAttention, we identify three key challenges.
\textit{(C1) Scale Explosion}: In tile-wise accumulation, approximation errors and scale changes from integer exponential operations propagate across tiles, causing numerical instability.
\textit{(C2) GPU Inefficiency}: The shift-based exponential approximation requires integer division to separate integer and fractional parts, which is significantly slower than multiplication or shift operations on GPUs.
\textit{(C3) Quantization Granularity}: Fused attention requires uniform scales across tiles for direct integer comparison; per-token quantization introduces scale mismatches that necessitate costly dequantization.

In this paper we propose QFlash, an integer-only fused attention on tile-based computation that addresses these challenges.
As shown in Figure~\ref{fig:1}, QFlash keeps the tile-based design but quantizes all operators including the softmax, thus the full kernel runs in the integer domain.
\textit{First}, we implement online softmax~\cite{milakov2018online} with shift-based exponential approximation and row-wise max updates, addressing C1 by carefully managing scale propagation across tiles.
\textit{Second}, we optimize the shift-based exponential computation to minimize GPU inefficiency (C2) through efficient integer arithmetic.
\textit{Third}, we adopt per-tensor quantization granularity to maintain uniform scales across the entire attention computation, enabling direct integer comparison without dequantization (C3).
This full kernel fusion reduces off-chip memory use, improves scheduling, and raises GPU utilization.
QFlash therefore keeps the benefits of fused attention while providing a fully integer path that improves both latency and energy efficiency.

To validate the proposed method, we implement QFlash as an integer-only kernel in Triton and evaluate it on an NVIDIA GeForce RTX 5090 with TVM 0.8, PyTorch 2.7.1 with CUDA 12.8, and Triton 3.3.1.
We extract seven representative attention workloads (A1--A7) from ViT, DeiT, and Swin models at $224\times224$ resolution to systematically evaluate latency performance.
On DeiT and ViT workloads (A1--A3), QFlash achieves up to $4.54\times$ speedup at batch 1 and $6.73\times$ at batch 8 compared with I-ViT.
On Swin workloads (A4--A7), the speedup reaches $7.69\times$ and $8.69\times$ respectively.
We also observe that QFlash's IMMA utilization is 8.0\%, lower than FlashAttention-2's HMMA utilization of 13.08\%, but since IMMA provides twice the peak throughput on RTX 5090, QFlash delivers higher absolute performance.

For the same workload, QFlash reduces energy consumption by 18.8\%, from 929.6$\mu$J with FP16 FlashAttention-2 to 754.6$\mu$J.
In terms of accuracy, QFlash improves SQNR by up to 6.7dB over I-ViT.
It also maintains Top-1 accuracy close to FP32 on ViT and DeiT models evaluated on ImageNet-1K.
These results show that an integer-only design can achieve high speed and efficiency without sacrificing accuracy in real inference.

\section{Background \& Related Work}

\subsection{FlashAttention}

Standard self-attention requires computing the similarity between all query key pairs and storing the results in a matrix, leading to memory and computational complexity of $O(N^2)$ with respect to the sequence length.

As a result, both the computation cost and GPU memory usage increase rapidly when processing long sequences.
FlashAttention was proposed to address this issue by leveraging the GPU memory hierarchy to avoid storing the intermediate similarity matrix in memory and instead performing computations in a tile-based manner.
The key idea is to use online softmax~\cite{milakov2018online}, which incrementally updates the softmax normalization as each block is processed.
In standard softmax, all similarity values must be collected at once to compute the denominator.

In contrast, online softmax processes similarity values block by block, updating the running maximum and exponential sum as new blocks are computed.
This approach maintains numerical stability while eliminating the need to store the entire similarity matrix in memory.
Thanks to this design, FlashAttention performs computations directly in high-speed GPU memory and writes results to global memory only when necessary.
As a result, the number of memory accesses is significantly reduced, and the overall memory complexity is relaxed from $O(N^2)$ to $O(N)$.

\subsection{Vision Transformer Quantization}

Prior work on Vision Transformer (ViT) compression has predominantly targeted linear operators, notably matrix multiplications~\cite{repq-vit,ptq4vit,zhong2024erq,clamp-vit}.
Subsequent studies extended quantization to non-linear components (Softmax, GELU, and LayerNorm) using polynomial, log-domain, or bit-shift approximations~\cite{kim2021bert,fq-vit,you2024shiftaddllm,illm,mixed_v1}.
These contributions, however, remain at the algorithmic level and do not translate the approximations into kernel-level speedups.

I-ViT~\cite{li2023vit} advances the state of the art by bit-shift approximating Softmax and GELU and applying TVM-based intra-operator optimizations.  
Nevertheless, it forgoes inter-operator fusions akin to FlashAttention, leaving memory traffic largely unoptimized and thereby preserving considerable headroom for further latency reduction in quantized ViTs.

There have been some efforts to optimize quantized attention kernels by fusing attention operators~\cite{kluska2024qattn,chen2024int}.
These approaches reduce computation and memory usage by quantizing the key matrix multiplications in attention.
However, the core operation of attention—Softmax—is still computed in the floating-point domain, which limits their applicability for integer-only inference.

More recent works accelerate attention and nonlinear operators through hardware–software co-design. 
Fused Tensor Core~\cite{jahadi2025fused} fuses matrix multiplication and Softmax on GPUs by offloading max and sum operations to Tensor Cores, but still relies on floating-point exponentiation. PICACHU~\cite{qin2025picachu} and QUARK~\cite{zhao2025quark} accelerate Softmax, GELU, and LayerNorm using dedicated hardware or CGRA-based designs.
While effective, these approaches either keep Softmax in the floating-point domain or depend on specialized hardware support, limiting their generality for software-only, integer-only attention on commodity GPUs.

To address these limitations, this work proposes a fully integer-only FlashAttention kernel that quantizes all attention operations, including Softmax, using a software-only design on commodity GPUs.

\subsection{Advantages of Integer-only Quantization}

\textbf{Peak performance of CUDA cores and Tensor Cores.}
For CUDA cores, the latency of integer and floating-point operations is almost the same, so full quantization is not very useful~\cite{arafa2019instructions}.
On the RTX 5090, however, measurements show that Tensor Cores can run 174K operators per cycle with HMMA (FP16) and 348K with IMMA (INT8), which is two times higher\footnote{Calculated based on 170 SMs configuration of RTX 5090 architecture.}.
Applying quantization to all operators therefore increases IMMA use and reduces data communication, giving a clear benefit.

\textbf{Performance per watt of Tensor Cores.}
Half-precision matrix multiply and accumulate (HMMA) uses less energy per operation, but the instruction overhead is still high, which reduces the overall performance per watt.
Integer matrix multiply (IMMA) may consume a bit more energy per operation, but the instruction overhead is lower, so it reaches higher compute density in real workloads~\cite{dally2023hardware}.  
From this view, quantizing the model to use IMMA improves both energy efficiency and performance density.
In this work, we show that using IMMA with quantization improves not only latency and throughput but also energy efficiency.

\textbf{Use of integer-only accelerators.}
In mixed-precision designs, matrix multiplications are quantized, but key operations such as softmax still depend on floating-point units.
This causes extra communication cost between integer and floating-point hardware, which limits the acceleration and adds inefficiency from floating-point units.
Low-cost integer-only accelerators cannot support such mixed-precision needs, so extra floating-point units must be added in a heterogeneous chip.
This increases chip area and power budget, which raises the cost of model deployment.

\begin{figure*}[!t]
\centering
\includegraphics[width=\textwidth]{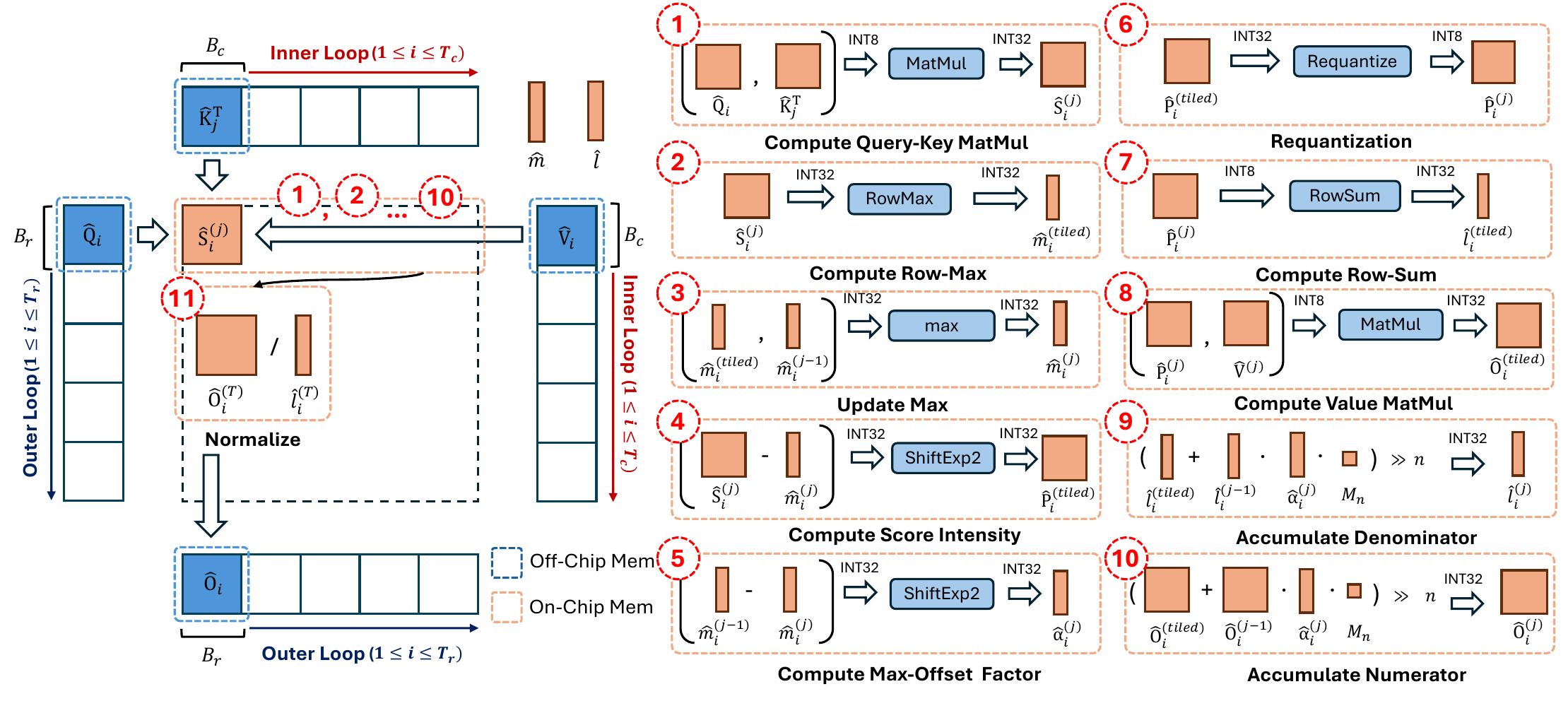}
\caption{Proposed QFlash forward pass (Algorithm~1), showing the integer-based attention pipeline with quantization, shift-exp softmax approximation, and blockwise accumulation.}
\label{fig:2}
\end{figure*}

\input{algorithm}

\section{Challenges of Fully Integer FlashAttention}

\subsection{C1: Scale Explosion in Tile-wise Integer Accumulation}

The first challenge is how to manage scaling in a tile-wise accumulation setting to avoid overflow and severe accuracy degradation. FlashAttention computes attention outputs tile by tile and incrementally accumulates each partial result into the running output.
Since the accumulated result from previous tiles is reused as input for subsequent tiles, numerical stability becomes critical once quantization is applied.
In integer-based FlashAttention, the exponential function is replaced with an integer approximation.
The resulting approximation errors and scale variations are repeatedly injected into the accumulation process across tiles, which can lead to rapid scale growth and instability if not carefully controlled.
A detailed comparison of scale management strategies is provided in Appendix~\ref{app:kernel_c1}.

\subsection{C2: Inefficiency of Shift-based Exp/Softmax on GPU}

The integer softmax proposed in I-ViT~\cite{li2023vit} typically transforms the exponential function into an $\exp2(\cdot)$ form and implements it using shift operations combined with linear approximations. While this approach is theoretically efficient, it introduces practical inefficiencies on GPUs when applied to integer inputs.

Specifically, computing $\exp2(x)$ requires decomposing the input into integer and fractional parts.
For quantized integer inputs, this decomposition involves integer division.
On GPUs, integer division has significantly higher latency than multiplication or shift operations,
making it a major performance bottleneck for softmax computation.
Implementation details of our GPU-friendly solution are discussed in Appendix~\ref{app:kernel_c2}.

\subsection{C3: Quantization Granularity in Integer Fused Attention}

Attention inputs have a multi-dimensional structure, and quantization granularity can be applied at the tensor, head, or token level.
The choice of granularity directly affects both accuracy and performance.

In FlashAttention, stable exponential computation requires subtracting the maximum value from $QK^\top$ scores,
with the maximum being updated at every tile.
When this process is performed in an integer-only setting, updating the maximum requires either
all tiles to be computed under the same scale or the values to be dequantized into floating point for comparison.
However, dequantization introduces additional computation and memory accesses,
which significantly undermines the performance benefits of fused attention kernels.

For this reason, we aim to maintain a shared scale that enables direct comparison of integer values without dequantization.
From this perspective, per-token quantization generates different scales for each token,
making it difficult to apply in integer fused attention during max updates and accumulation.
As a result, per-tensor and per-head quantization are the practical granularity choices for integer-based FlashAttention.
A visual illustration of granularity constraints is presented in Appendix~\ref{app:kernel_c3}.

\section{Integer-Only Fused Attention Method}

In this section, we describe in detail how the proposed QFlash algorithm operates.
We first define the notation used in the description.
Let $b$ denote the bit-width, and let $\mathbb{I}_{\mathrm{int}(b)}$ denote the set of signed $b$-bit integers as defined in Equation~\eqref{eq:basic_quant}.
Let $s_X$ be the scale factor, and let $\hat{\mathbf{X}}$ denote the integer quantization of $\mathbf{X}$.

\begin{equation}
\mathbb{I}_{\mathrm{int}(b)} := \{ z \in \mathbb{Z} \mid -2^{b-1} \leq z \leq 2^{b-1}-1 \}.
\label{eq:basic_quant}
\end{equation}

For a real matrix $\mathbf{X} \in \mathbb{R}^{m \times n}$, the scale factor $s_X$ and the quantized matrix $\hat{\mathbf{X}}$ are defined as in Equation~\eqref{eq:basic_quant_X}.

\begin{equation}
s_X = \frac{\|\mathbf{X}\|_\infty}{2^{b-1}-1} \in \mathbb{R}, \quad
\hat{\mathbf{X}} = \left\lfloor \frac{\mathbf{X}}{s_X} \right\rceil
\in \mathbb{I}^{m \times n}_{\mathrm{int}(b)}.
\label{eq:basic_quant_X}
\end{equation}

QFlash maintains the tile-based fused attention computation of FlashAttention~\cite{dao2022flashattention} while quantizing all operators, including the softmax, to achieve both data size reduction and compute efficiency.
The detailed procedure follows Figure~\ref{fig:2} and Algorithm~\ref{algo:1}.
One execution of the outer loop in Algorithm~\ref{algo:1} computes a single output tensor $\hat{\mathbf{O}}_i$.
As shown in Figure~\ref{fig:2}, the process consists of the following eleven steps:
\circnum{1} Compute Query--Key Score $\rightarrow$ \circnum{2} Compute Row-Max $\rightarrow$ \circnum{3} Update Max $\rightarrow$ \circnum{4} Compute Score Intensity $\rightarrow$ \circnum{5} Compute Max-Offset Factor $\rightarrow$ \circnum{6} Requantization $\rightarrow$ \circnum{7} Compute Row-Sum $\rightarrow$ \circnum{8} Compute Value MatMul $\rightarrow$ \circnum{9} Accumulate Denominator $\rightarrow$ \circnum{10} Accumulate Numerator $\rightarrow$ \circnum{11} Normalize.

\subsection{MatMul}

QFlash performs integer matrix multiplications for \circnum{1} Compute Query--Key MatMul and \circnum{8} Compute Value MatMul.
These operations can be expressed as the dot product of two quantized matrices $\hat{\mathbf{A}} \in \mathbb{I}^{B_r \times d}_{\text{int}8}$ and $\hat{\mathbf{B}} \in \mathbb{I}^{d \times B_c}_{\text{int}8}$.
Here, $B_r$ denotes the size of the query block, $B_c$ denotes the size of the key/value block, and $d$ represents the inner dimension of the multiplication.
The inputs are given in int8 format, and the results are accumulated in int32 format.
Equation~(3) defines the scaling relationship corresponding to this integer matrix multiplication.

\begin{equation}
\hat{\mathbf{C}} = \hat{\mathbf{A}} \cdot \hat{\mathbf{B}}
\in \mathbb{I}_{\text{int}32}^{B_r \times B_c},
\quad
s_C = s_A \cdot s_B \in \mathbb{R}.
\end{equation}

This operation can be highly optimized on GPUs and dedicated accelerators, replacing floating-point multiplications to significantly reduce both computational cost and memory access overhead.

\subsection{Compute Row-Max and Update Max}

For a matrix $\hat{\mathbf{X}} \in \mathbb{I}^{B_r \times B_c }_{\text{int}32}$, the \circnum{2} Compute Row-Max step calculates the maximum value of each row to obtain a vector $\hat{m} \in \mathbb{I}_{\mathrm{int}32}^{B_r}$.
Subsequently, in the \circnum{3} Update Max step, these values are compared with those from the previous tile to update the row-wise maxima progressively.

\begin{equation}
\hat{m}_i = \max_{0 \le j < B_c} \hat{\mathbf{X}}[i, j],
\quad
\hat{m} = \big[\hat{m}_1, \dots, \hat{m}_{B_r}\big] \in \mathbb{I}_{\mathrm{int}32}^{B_r}.
\end{equation}

Equation~(4) defines the extraction of the maximum value from each row, which serves as the reference point for the subsequent softmax computation.
By subtracting the row-wise maximum from all elements, potential overflow in the exponential calculation is prevented.
In addition, \textit{Update Max} incorporates the running maximum across tiles during block-wise operations, ensuring stable numerical computation throughout the entire sequence.
This tile-wise max update requires consistent scaling for direct integer comparison, which motivates Challenge~C3.

\subsection{Compute Score Intensity and Max-Offset Factor}

In QFlash, the exponential operation is used in step \circnum{4} Compute Score Intensity and step \circnum{5} Compute Max-Offset Factor.
The Score Intensity step computes the exponential term of softmax in integer form to measure attention strength.
The Max-Offset Factor step uses the Row-Max value to keep the exponential operation stable.

We adapt the ShiftExp algorithm from I-ViT to approximate the exponential function of softmax with integer arithmetic.
The key of softmax is the $e^x$ computation.
By multiplying with the constant $\log_2(e)$, we can rewrite it as a power of two.
The input matrix is $\hat{\mathbf{X}} \in \mathbb{I}_{\text{int}32}^{m \times n}$ with scale factor $s_X \in \mathbb{R}$.
After Row-Max is applied, $\hat{\mathbf{X}}$ always has negative values; therefore, the exponential is applied only to negative inputs.
This process is given in Equation~(5).

\begin{equation}
\scalebox{1.2}{$
e^{s_X \cdot \hat{\mathbf{X}}} =
2^{(s_X \cdot \log_2e )\cdot \hat{\mathbf{X}}} =
2^{\tilde{s}_X \cdot \hat{\mathbf{X}}},
$}
\end{equation}

Here, $\tilde{s}_X = s_X \cdot \log_2 e$.
Equation~(6) rewrites the input $\hat{\mathbf{X}}$ as an integer part $\mathbf{Q} \in \mathbb{Z}_{\leq 0}$ and a fractional part $\mathbf{R}$.
This decomposition typically requires integer division on GPUs, which motivates Challenge~C2.

\begin{equation}
\scalebox{1.2}{$
2^{\tilde{s}_X \cdot  \hat{\mathbf{X}}} =
2^{(\tilde{s}_X \cdot \mathbf{R}) + \mathbf{Q}}
$}
\end{equation}

Since the fractional part $\mathbf{R}$ lies in $(-1, 0]$, it can be replaced with a simple linear approximation as in Equation~(7).

\begin{equation}\
\scalebox{1.2}{$
2^{(\tilde{s}_X \cdot \mathbf{R})} \approx
\frac{\tilde{s}_X \cdot \mathbf{R}}{2} + 1 =
\tilde{s}_X \cdot \left(\frac{\mathbf{R}}{2}  + 
\left\lfloor \frac{1}{\tilde{s}_X} \right\rceil \right)
$}
\end{equation}

Note that $\left\lfloor \frac{1}{\tilde{s}_X} \right\rceil$ is a constant determined only by $\tilde{s}_X$.
Thus, it is precomputed outside the kernel, and the kernel performs only element-wise integer operations.
Finally, Equation~(8) shows the combination of the integer part and the approximated fractional part to compute the exponential.

\begin{equation}
\scalebox{1.2}{$
e^{s_X \cdot \hat{\mathbf{X}}} \approx
\tilde{s}_X \cdot \left(\frac{\mathbf{R}}{2} 
+ \left\lfloor \frac{1}{\tilde{s}_X} \right\rceil \right)  
\cdot 2^{\mathbf{Q}}
$}
\end{equation}

This formulation enables an efficient integer-only approximation of the exponential operation using simple arithmetic and bit-shift operations.

\subsection{Requantization}

The result of the matrix multiplication is stored as $\hat{\mathbf{X}} \in \mathbb{I}_{\text{int}32}^{B_r \times B_c}$.
To be used in subsequent operations, this value must be converted back into the range of $\hat{\mathbf{Y}} \in \mathbb{I}_{\text{int}8}^{B_r \times B_c}$ through the \circnum{6} Requantization step.
This is achieved by applying a scaling transformation that reduces the integer range.
First, we define a fixed-point multiplier as in Equation~(9).
Here, the scale ratio is converted into a binary logarithm to compute $n$, and the shift size $r$ is set by considering the bit-width $b$.

\begin{equation}
n = \left\lfloor \log_2\left(\frac{s_X}{s_Y}\right) \right\rfloor,
\quad
r = b - n.
\end{equation}

Based on this, the requantization is performed as shown in Equation~(10).

\begin{equation}
M_r = \left\lfloor \frac{s_X}{s_Y} \cdot 2^{r} \right\rceil, \quad
\hat{\mathbf{Y}}  =  \left ( \hat{\mathbf{X}} \cdot 
M_r \right )
\gg r.
\end{equation}

We precompute $n$, $r$, and $\left\lfloor \frac{s_X}{s_Y} \cdot 2^{r} \right\rceil$ outside the kernel, since they depend only on the scale factors and bit-width and are independent of the input $\hat{\mathbf{X}}$.

\subsection{Compute Row-Sum}

For softmax normalization, the denominator requires the sum of each row.
Thus, for the matrix $\hat{\mathbf{X}} \in \mathbb{I}^{B_r \times B_c}_{\text{int}8}$,
the \circnum{7} Compute Row-Sum step calculates the element-wise sum of each row.

\begin{equation}
\hat{s}_i = \sum_{j=1}^{B_c} \hat{\mathbf{X}}[i, j],
\quad
\hat{s} = \big[\hat{s}_1, \dots, \hat{s}_{B_r}\big] \in \mathbb{I}_{\text{int}8}^{B_r}.
\end{equation}

Equation~(11) describes the computation of the row-wise sums,
where $\hat{s}_i$ represents the sum of the elements in the $i$-th row.
These values serve as the key denominator terms in the softmax normalization.

\subsection{Accumulate Numerator and Denominator}

For softmax normalization, steps \circnum{9} Accumulate Denominator and \circnum{10} Accumulate Numerator must be performed.
If these accumulations are carried out directly in the quantized domain, overflow may occur.
A common approach dequantizes values before accumulation, but this introduces floating-point operations and weakens integer-only benefits.
Related to Challenge~C1, we approximate the inverse scale as an integer to perform division, enabling stable numerator/denominator accumulation and integer-only softmax normalization without overflow.
We refer to this operation as $\operatorname{ScaleRelease}$, whose definition and analysis are provided in Appendix~\ref{app:kernel_c1}.

\subsection{Normalization}

After the inner loop, step \circnum{11} performs \textit{normalization}.
Both the numerator and denominator are accumulated as integers, and their ratio corresponds to softmax normalization without explicitly materializing the full score matrix.
Since the scale factors cancel out during normalization, dequantization is implicitly handled.
Importantly, we implement this division \textit{entirely in integer arithmetic}, eliminating floating-point operations while producing int8 outputs directly usable for downstream computation.

\section{Evaluation}

All experiments were run on an NVIDIA GeForce RTX 5090 GPU with TVM 0.8, PyTorch 2.7.1 (CUDA 12.8), and Triton 3.3.1. We evaluated latency, energy, and accuracy at the operator level and then tested Top-1 accuracy against I-ViT, I-BERT, FQ-ViT, FlashAttention-2, INT-FlashAttention (Full/Half), and QAttn.

We extract seven unique attention workloads from different ViT/DeiT/Swin model variants, as summarized in Table~\ref{tab:attn_workloads_vertical}.
In addition, we evaluate both batch size 1 and batch size 8 to cover real-time inference and throughput-oriented inference settings, respectively.

\begin{table}[t]
  \scriptsize
  \centering
  \caption{Unique attention workload configurations used in ViT/DeiT and Swin inference at $224\times224$.
  $\#Win$ denotes the number of windows (only applicable to Swin), $H$ the number of heads, $N$ the context length, and $D$ the head dimension.}
  \label{tab:attn_workloads_vertical}
  \renewcommand{\arraystretch}{1.15}
  \setlength{\tabcolsep}{4pt}

  \begin{tabular}{@{} c l c c c c l @{}}
    \toprule
    \textbf{Workload} & \textbf{Source} & \textbf{\#Win} & \textbf{$H$} & \textbf{$N$} & \textbf{$D$} & \textbf{Shape ($W{\times}H{\times}N{\times}D$)} \\
    \midrule

    A1 & ViT/DeiT-Tiny  & \redx & 3  & 197 & 64 & $(3 \times 197 \times 64)$ \\
    A2 & ViT/DeiT-Small & \redx & 6  & 197 & 64 & $(6 \times 197 \times 64)$ \\
    A3 & ViT/DeiT-Base  & \redx & 12 & 197 & 64 & $(12 \times 197 \times 64)$ \\
    \midrule
    A4 & Swin-T/S Stage-1 & 64 & 3  & 49 & 32 & $(64 \times 3 \times 49 \times 32)$ \\
    A5 & Swin-T/S Stage-2 & 16 & 6  & 49 & 32 & $(16 \times 6 \times 49 \times 32)$ \\
    A6 & Swin-T/S Stage-3 & 4  & 12 & 49 & 32 & $(4 \times 12 \times 49 \times 32)$ \\
    A7 & Swin-T/S Stage-4 & 1  & 24 & 49 & 32 & $(1 \times 24 \times 49 \times 32)$ \\

    \bottomrule
  \end{tabular}
\end{table}

\subsection{Inference Latency}

\begin{figure}[t]
\centering
\begin{subfigure}[t]{0.48\textwidth}
  \centering
  \includegraphics[width=\textwidth]{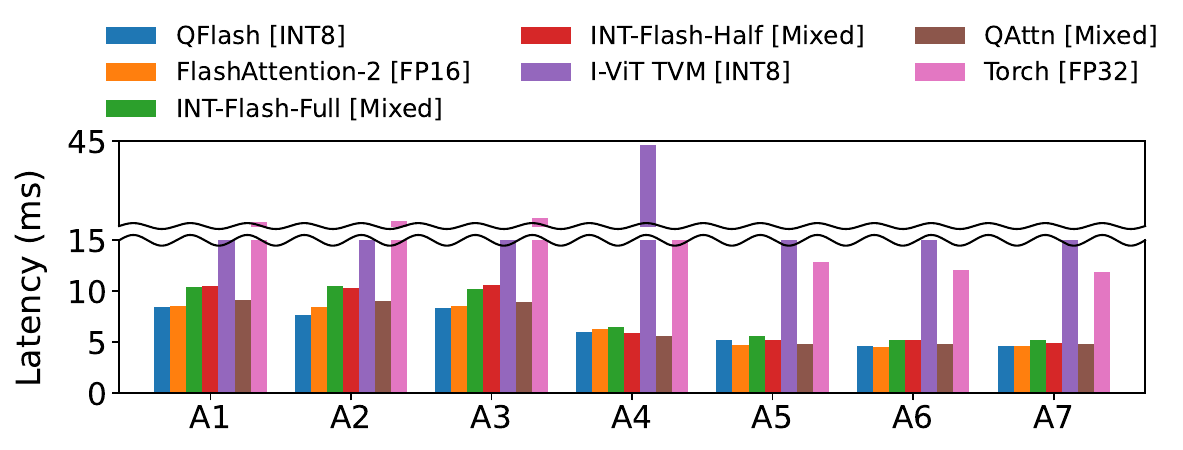}
  \caption{}
  \label{fig:latency_batch1}
\end{subfigure}
\hfill
\begin{subfigure}[t]{0.48\textwidth}
  \centering
  \includegraphics[width=\textwidth]{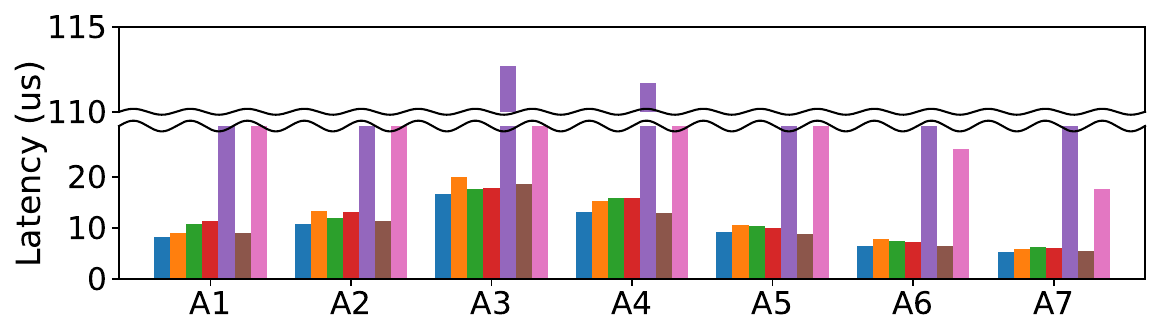}
  \caption{}
  \label{fig:latency_batch8}
\end{subfigure}
\caption{Latency comparison of different attention kernels across the seven attention workloads (A1--A7) listed in Table~\ref{tab:attn_workloads_vertical}.
Results are reported for (a) batch size 1 and (b) batch size 8.}
\label{fig:3}
\end{figure}

\noindent\textbf{Latency comparison.}
Figure~\ref{fig:3} compares the latency of different attention kernels across the seven attention workloads listed in Table~\ref{tab:attn_workloads_vertical}.
Since I-ViT is a representative \emph{integer-only} attention kernel, we measure speedup primarily against I-ViT to assess the effectiveness of our fully integer QFlash design.
In the batch-1 case, QFlash achieves up to $4.54\times$ and $7.69\times$ speedup over I-ViT on ViT/DeiT and Swin workloads, respectively.
In the batch-8 case, QFlash is up to $6.73\times$ and $8.69\times$ faster than I-ViT on ViT/DeiT and Swin workloads, respectively.
Despite the overhead of fully integer execution, QFlash consistently outperforms the integer-only baseline and remains generally faster or comparable to FP16 and mixed-precision kernels across workloads.

\noindent\textbf{Tensor Core utilization.}
To analyze the latency improvement of our kernel code, we used NVIDIA Nsight Compute~\cite{leinhauser2021performance} to profile Tensor Core utilization.
We report utilization as a ratio to the peak throughput, using two workloads (A2 and A7) with batch size 8.
The results are summarized in Table~\ref{table:tc_utilization}.
The key observation is that on the RTX 5090, Tensor Cores achieve 174K operators per cycle with HMMA (FP16) and 348K with IMMA (INT8), meaning the peak throughput of IMMA is exactly twice that of HMMA.
Therefore, even though the IMMA utilization of QFlash is only 8.0\% compared to 13.08\% HMMA utilization in FlashAttention-2, the higher peak throughput of IMMA means that QFlash still delivers greater overall performance.
The same conclusion holds for A7.

\begin{table}[t]
  \scriptsize
  \centering
  \caption{Tensor Core utilization (Peak \%) for different attention implementations on workloads A2 and A7 with batch size 8.}
  \label{table:tc_utilization}
  \renewcommand{\arraystretch}{1.15}
  \setlength{\tabcolsep}{4pt}

  \begin{tabular}{@{} l c c c c @{}}
    \toprule
    \multirow{2}{*}{\raisebox{-0.8ex}{\textbf{Method}}}
      & \multicolumn{2}{c}{\textbf{A2}}
      & \multicolumn{2}{c}{\textbf{A7}} \\
    \cmidrule(lr){2-3} \cmidrule(lr){4-5}
      & \textbf{IMMA} & \textbf{HMMA} & \textbf{IMMA} & \textbf{HMMA} \\
    \midrule

    Flash-2        & --     & 13.08\% & --     & 4.69\% \\
    QAttn          & 3.75\% & 7.51\%  & 1.25\% & 2.51\% \\
    INT-Flash-Half & 3.38\% & 6.75\%  & 1.12\% & 2.25\% \\
    INT-Flash-Full & 7.18\% & --      & 2.13\% & --     \\

    \rowcolor{oursrow}
    \textbf{QFlash (Ours)} & 8.00\% & -- & 2.55\% & -- \\

    \bottomrule
  \end{tabular}
\end{table}

\subsection{Energy Consumption}

We compared the energy consumption of five kernels on workload A2 with batch size 8.
Power consumption was recorded using \texttt{nvidia-smi}~\cite{yu2023know,bridges2016understanding},
and the number of operations was measured with Nsight Compute.
All kernels executed the same total amount of computation.
As listed in Table~\ref{table:energy_a2_b8}, QFlash shows the lowest energy consumption.
There are two reasons for this result.
First, QFlash achieves shorter execution time for the same workload, which reduces the integrated energy.
Second, QFlash consistently uses the IMMA instruction for Tensor Cores, which provides higher energy efficiency per operation compared to HMMA~\cite{dally2023hardware,horowitz20141}.
The combination of these two effects makes QFlash the kernel with the lowest energy consumption under identical conditions.

\begin{table}[t]
  \scriptsize
  \centering
  \caption{Energy consumption ($\mu$J) on workload A2 (Batch = 8).}
  \label{table:energy_a2_b8}
  \renewcommand{\arraystretch}{1.15}
  \setlength{\tabcolsep}{4pt}

  \begin{tabular}{@{} l c c c c c @{}}
    \toprule
    \textbf{Workload}
      & \textbf{Flash}
      & \textbf{INT-Flash (Full)}
      & \textbf{INT-Flash (Half)}
      & \textbf{QAttn}
      & \textbf{QFlash (Ours)} \\
    \midrule

    \textbf{A2 (B=8)} 
      & 929.6
      & 840.0
      & 920.5
      & 795.2
      & \textbf{754.6} \\
    \bottomrule
  \end{tabular}
\end{table}

\subsection{Accuracy Evaluation}

\begin{table}[t]
  \scriptsize
  \centering
  \caption{Quantitative evaluation of attention implementations in terms of SQNR and MSE.}
  \label{table:1}
  \renewcommand{\arraystretch}{1.15}
  \setlength{\tabcolsep}{4pt}

  \begin{tabular}{@{} l c c c c @{}}
    \toprule
    \multirow{2}{*}{\raisebox{-1.0ex}{\textbf{Method}}}
      & \multicolumn{2}{c}{\textbf{A2}}
      & \multicolumn{2}{c}{\textbf{A7}} \\
    \cmidrule(lr){2-3} \cmidrule(lr){4-5}
      & \textbf{SQNR} & \textbf{MSE} & \textbf{SQNR} & \textbf{MSE} \\
    \midrule

    I-ViT          & 25.80 & 7.05e-3 & 25.22 & 9.38e-3 \\
    QAttn          & 34.12 & 1.04e-3 & 34.75 & 1.04e-3 \\
    INT-Flash-Half & 38.19 & 4.04e-4 & 39.07 & 3.86e-4 \\
    INT-Flash-Full & 36.92 & 5.41e-4 & 37.80 & 5.17e-4 \\

    \rowcolor{oursrow}
    \textbf{QFlash (Ours)} & 32.50 & 1.51e-3 & 31.02 & 2.47e-3 \\

    \bottomrule
  \end{tabular}
\end{table}

\begin{table}[t]
  \scriptsize
  \centering 
  \caption{Accuracy and efficiency comparison across quantization granularities.
Symbols indicate granularity: $\triangle$~per-token, $\circ$~per-head, $\triangledown$~per-tensor.
For fairness, we use each prior work’s original algorithm and kernel implementation, which are tailored to their target granularity.}
  \label{table:accuracy_efficiency}
  \renewcommand{\arraystretch}{0.8}
  \setlength{\tabcolsep}{4pt}

  \begin{tabular}{@{} l l c c c c c c @{}}
    \toprule
    \textbf{Model (FP32 Acc)} & \textbf{Method} & \textbf{BOPs (G)} & \textbf{Int-Only} & \textbf{Acc} & \textbf{Q} & \textbf{K} & \textbf{V} \\
    \midrule

    \rowcolor{oursrow}
    \multirow{7}{*}{ViT-S (FP32: 81.38)}   
      & QFlash         & 17.2  & \greencheck & \textbf{82.24} & \qtensor   & \qtensor   & \qtensor   \\
      & I-ViT          & 17.2  & \greencheck & 81.19          & \qtensor & \qtensor & \qtensor \\
      & I-BERT          & 17.2  & \greencheck       & 81.00          & \qtensor & \qtensor & \qtensor \\
      & FQ-ViT          & 17.2  & \greencheck       & 81.06          & \qtensor & \qtensor & \qtensor \\
      & INT-Flash-Full & 17.2  & \redx       & 80.60          & \qtoken  & \qtoken  & \qhead   \\
      & INT-Flash-Half & 34.3  & \redx       & 80.62          & \qtoken  & \qtoken  & {--}     \\
      & QAttn          & 34.3  & \redx       & 80.23          & \qtensor & \qtensor & \qtensor \\
      
    \midrule
    
    \rowcolor{oursrow}
    \multirow{7}{*}{ViT-B (FP32: 85.10)}  
      & QFlash         & 34.3  & \greencheck & \textbf{86.84} & \qtensor   & \qtensor   & \qtensor   \\
      & I-ViT          & 34.3  & \greencheck & 85.02          & \qtensor & \qtensor & \qtensor \\
      & I-BERT          & 34.3  & \greencheck       & 83.97          & \qtensor & \qtensor & \qtensor \\
      & FQ-ViT          & 34.3  & \greencheck       & 84.82          & \qtensor & \qtensor & \qtensor \\
      & INT-Flash-Full & 34.3  & \redx       & 84.74          & \qtoken  & \qtoken  & \qhead   \\
      & INT-Flash-Half & 68.7  & \redx       & 84.83          & \qtoken  & \qtoken  & {--}     \\
      & QAttn          & 68.7  & \redx       & 84.68          & \qtensor & \qtensor & \qtensor \\
    \midrule

    \rowcolor{oursrow}
    \multirow{7}{*}{DeiT-T (FP32: 72.21)}
      & QFlash         & 17.2  & \greencheck & \textbf{71.70} & \qtensor   & \qtensor   & \qtensor   \\
      & I-ViT          & 17.2  & \greencheck & 71.70          & \qtensor & \qtensor & \qtensor \\
      & I-BERT           & 17.2  & \greencheck       & 71.20          & \qtensor & \qtensor & \qtensor \\
      & FQ-ViT          & 17.2  & \greencheck       & 71.53          & \qtensor & \qtensor & \qtensor \\
      & INT-Flash-Full & 17.2  & \redx       & 71.64          & \qtoken  & \qtoken  & \qhead   \\
      & INT-Flash-Half & 34.3  & \redx       & 71.64          & \qtoken  & \qtoken  & {--}     \\
      & QAttn          & 34.3  & \redx       & 71.25          & \qtensor & \qtensor & \qtensor \\
    \midrule

    \rowcolor{oursrow}
    \multirow{7}{*}{DeiT-S (FP32: 79.85)}
      & QFlash         & 34.3  & \greencheck & 79.46          & \qtensor   & \qtensor   & \qtensor   \\
      & I-ViT          & 34.3  & \greencheck & 79.53          & \qtensor & \qtensor & \qtensor \\
      & I-BERT          & 34.3  & \greencheck       & 79.48          & \qtensor & \qtensor & \qtensor \\
      & FQ-ViT          & 34.3  & \greencheck       & 79.39          & \qtensor & \qtensor & \qtensor \\
      & INT-Flash-Full & 34.3  & \redx       & 79.64          & \qtoken  & \qtoken  & \qhead   \\
      & INT-Flash-Half & 68.7  & \redx       & \textbf{79.68} & \qtoken  & \qtoken  & {--}     \\
      & QAttn          & 68.7  & \redx       & 79.43          & \qtensor & \qtensor & \qtensor \\
    \midrule

    \rowcolor{oursrow}
    \multirow{7}{*}{DeiT-B (FP32: 81.85)}
      & QFlash         & 68.7  & \greencheck & 81.59          & \qtensor   & \qtensor   & \qtensor   \\
      & I-ViT          & 68.7  & \greencheck & 81.47          & \qtensor & \qtensor & \qtensor \\
      & I-BERT          & 68.7 & \greencheck       & 81.59          & \qtensor & \qtensor & \qtensor \\
      & FQ-ViT          & 68.7 & \greencheck       & 81.72          & \qtensor & \qtensor & \qtensor \\
      & INT-Flash-Full & 68.7  & \redx       & 81.85          & \qtoken  & \qtoken  & \qhead   \\
      & INT-Flash-Half & 137.0 & \redx       & \textbf{81.90} & \qtoken  & \qtoken  & {--}     \\
      & QAttn          & 137.0 & \redx       & 81.85          & \qtensor & \qtensor & \qtensor \\
    \midrule

    \rowcolor{oursrow}
    \multirow{7}{*}{Swin-T (FP32: 81.35)}
      & QFlash         & 13.5  & \greencheck & 80.06          & \qtensor   & \qtensor   & \qtensor   \\
      & I-ViT          & 13.5  & \greencheck & 80.83 & \qtensor & \qtensor & \qtensor \\
      & I-BERT          & 13.5   & \greencheck       & \textbf{81.11}          & \qtensor & \qtensor & \qtensor \\
      & FQ-ViT         & 13.5   & \greencheck       & 80.90          & \qtensor & \qtensor & \qtensor \\
      & INT-Flash-Full & 13.5  & \redx       & 80.04          & \qtoken  & \qtoken  & \qhead   \\
      & INT-Flash-Half & 26.9  & \redx       & 80.05          & \qtoken  & \qtoken  & {--}     \\
      & QAttn          & 26.9  & \redx       & 80.13          & \qtensor & \qtensor & \qtensor \\
    \midrule

    \rowcolor{oursrow}
    \multirow{7}{*}{Swin-S (FP32: 83.20)}
      & QFlash         & 22.0  & \greencheck & 81.86          & \qtensor   & \qtensor   & \qtensor   \\
      & I-ViT          & 22.0  & \greencheck & 82.81          & \qtensor & \qtensor & \qtensor \\
      & I-BERT          & 22.0  & \greencheck       & 83.09          & \qtensor & \qtensor & \qtensor \\
      & FQ-ViT          & 22.0  & \greencheck       & 83.15          & \qtensor & \qtensor & \qtensor \\
      & INT-Flash-Full & 22.0  & \redx       & 82.11          & \qtoken  & \qtoken  & \qhead   \\
      & INT-Flash-Half & 43.9  & \redx       & \textbf{83.20} & \qtoken  & \qtoken  & {--}     \\
      & QAttn          & 43.9  & \redx       & 82.24          & \qtensor & \qtensor & \qtensor \\
    \bottomrule
  \end{tabular}
\end{table}

\noindent\textbf{Quantization-level accuracy.}
Accuracy was measured with Signal-to-Quantization-Noise Ratio (SQNR), which indicates the relative signal-to-noise ratio after quantization, and Mean Squared Error (MSE), which quantifies the absolute error.
Table~\ref{table:1} reports the results for two attention workloads, A2 and A7, evaluated with batch size 8.
For both input sizes, QFlash shows SQNR above 30dB and MSE around $10^{-3}$,
which means the distortion from quantization is small.
The quality is higher than the baseline integer method I-ViT, but slightly lower than INT-Flash-Half and INT-Flash-Full due to their mixed-precision design.
QFlash chooses a design that favors speed and energy efficiency, while still giving acceptable accuracy and a balanced trade-off.

\noindent\textbf{Top-1 evaluation setup.}
We measure the end-to-end Top-1 accuracy by running ViT, DeiT, and Swin models with quantized attention applied during inference.
Specifically, we quantize only the attention modules while keeping the remaining layers in floating point to isolate the impact of attention quantization.
For activation scaling, we use dynamic quantization, where scaling factors are computed on-the-fly during inference.
Since I-ViT, I-BERT, FQ-ViT, INT-FlashAttention, and QAttn assume different quantization granularities, we use their original implementations without any modification for a fair comparison.
As shown in Table~\ref{table:accuracy_efficiency}, I-ViT, I-BERT, FQ-ViT, and QAttn use $\triangledown$~per-tensor quantization, while INT-FlashAttention uses $\triangle$~per-token and $\circ$~per-head quantization.
In general, finer granularity requires more scaling factors, which leads to a trade-off between accuracy and efficiency.

\noindent\textbf{Top-1 accuracy results.}
Top-1 classification accuracy is reported in Table~\ref{table:accuracy_efficiency}.
QFlash maintains accuracy close to FP32 on ViT and DeiT models, and achieves better accuracy than I-ViT, I-BERT, and FQ-ViT on ViT backbones.
However, on Swin models, QFlash shows lower Top-1 accuracy than I-BERT and FQ-ViT, since window partitioning is also parallelized and thus amplifies quantization error under $\triangledown$~per-tensor scaling.
These results indicate that QFlash effectively mitigates quantization error by stabilizing scaling factors within each tile, even under $\triangledown$~per-tensor quantization with only a small number of scaling factors.
Overall, QFlash enables an efficient integer-only attention kernel with minimal scaling overhead via per-tensor quantization.

\section{Conclusion}

In this paper, we proposed QFlash to address the dependence of Transformer attention on floating-point operations and the off-chip memory bottleneck from intermediate tensors.
QFlash implemented the full attention process, including softmax, using INT8/INT32 integer-only operations with shift-based log/exp approximations and integer normalization.
Experiments showed that QFlash achieved higher speed and precision than I-ViT and improved energy efficiency compared to FP16 FlashAttention, providing a practical solution for efficient and accurate integer-only attention in large-scale Transformer inference.

\bibliographystyle{named}
\bibliography{simple-base}

\clearpage

\appendix
\section*{Appendix}
\addcontentsline{toc}{section}{Appendix}

\section{Step-wise Application of Integer Operations to the Kernel}
\label{app:kernel_improvement}
\begin{figure}[H]
\centering
\includegraphics[width=0.5\textwidth]{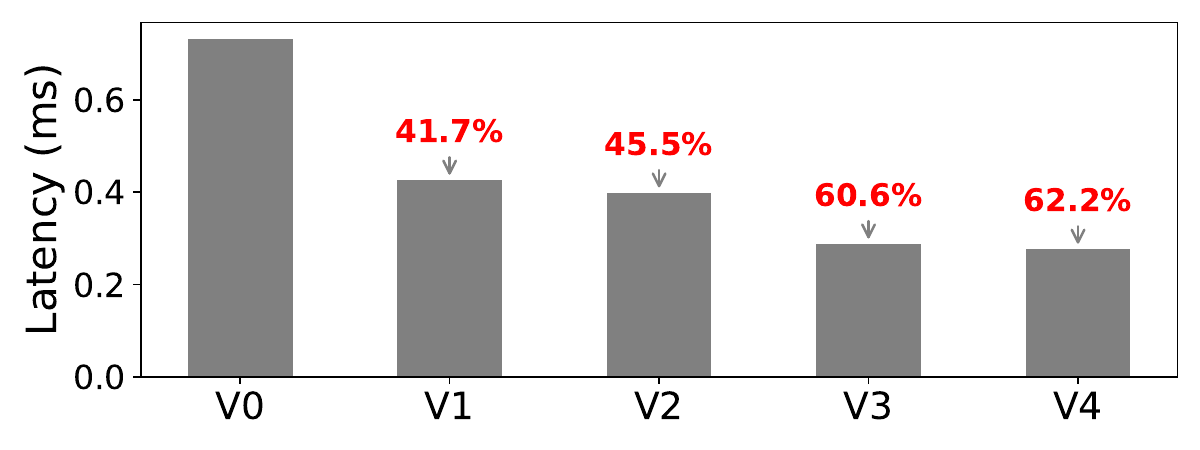}
\caption{Kernel optimization results}
\label{fig:4}
\end{figure}

Figure~\ref{fig:4} illustrates the speedup achieved by progressively applying integer operations to the FlashAttention-2 kernel.
The experiment is conducted on the A2 workload with a batch size of 1024.
V0 is the baseline FlashAttention-2 kernel.
V1 replaces the QK matrix multiplication with an INT8 GEMM, yielding a 41.4\% speedup.
V2 further replaces both QK and PV matrix multiplications with INT8 GEMMs, achieving a 46.0\% speedup.
V3 additionally applies an integer-approximated exponential operation, resulting in a 61.0\% speedup.
Finally, V4 applies integer operations to QK, PV, the approximated exponential, and tile-wise accumulation, achieving up to a 62.4\% speedup.

\section{Detailed Analysis of QFlash Challenges}
\subsection{C1: Extended Analysis of Tile-wise Accumulation under Integer Quantization}
\label{app:kernel_c1}

FlashAttention computes attention outputs in a tile-wise manner and progressively accumulates partial results from each tile into the running output. Since the output from previous tiles is reused as input for subsequent tiles, maintaining numerical stability throughout the accumulation process is critical.
In integer-quantized FlashAttention, the exponential function is replaced with an integer approximation, and the approximation error can repeatedly propagate during tile-wise accumulation. Therefore, the key challenge is how to manage scaling factors in the tile-based accumulation scheme.

\subsubsection{Comparison between Scale Accumulation and Scale Release}

For each query position $i$, $\mathbf{O}_i^{(j-1)}$ denotes the accumulated attention output computed up to tile $j-1$. To ensure numerical stability in the numerator accumulation of FlashAttention, the running maximum is updated from $m_i^{(j-1)}$ to $m_i^{(j)}$, and the scale of the previously accumulated output is adjusted accordingly. 
Specifically, we define $\alpha_i^{(j)} = \exp\left(m_i^{(j-1)} - m_i^{(j)}\right)$, rescale the previous output $\mathbf{O}_i^{(j-1)}$ by multiplying $\alpha_i^{(j)}$, and then add the current tile contribution $\mathbf{P}_i^{(j)}\mathbf{V}_j$ to obtain the updated accumulated output $\mathbf{O}_i^{(j)}$.

\begin{equation}
\mathbf{O}_i^{(j)} \gets \mathbf{O}_i^{(j-1)} \cdot \alpha_i^{(j)} + \mathbf{P}_i^{(j)} \mathbf{V}_j
\end{equation}

If we perform integer accumulation and approximate the exponential function with integer arithmetic, we can choose one of the following two approaches.

The \textbf{Scale Accumulation} approach continuously accumulates the scale of $\hat{\mathbf{O}}_i^{(j)}$.

\begin{equation}
\hat{\mathbf{O}}_i^{(j)} \gets \hat{\mathbf{O}}_i^{(j-1)} \cdot \hat{\alpha}_i^{(j)} +
\left\lfloor \frac{\hat{\mathbf{P}}_i^{(j)} \hat{\mathbf{V}}_j}{s_\alpha} \right\rfloor
\end{equation}

This approach works reliably for operator-wise independent integer kernels. 
However, in FlashAttention, where tile-wise accumulation is repeatedly performed, the value range grows rapidly as the number of accumulation steps increases, leading to overflow and severe accuracy degradation.

The \textbf{Scale Release} approach releases the scale of the exponential term at each accumulation step, keeping the scale of $\hat{\mathbf{O}}_i^{(j)}$ constant.

\begin{equation}
\hat{\mathbf{O}}_i^{(j)} \gets \lfloor \hat{\mathbf{O}}_i^{(j-1)} \cdot \hat{\alpha}_i^{(j)} \cdot s_\alpha \rfloor + \hat{\mathbf{P}}_i^{(j)} \hat{\mathbf{V}}_j
\end{equation}

\begin{figure}[H]
\centering
\includegraphics[width=0.5\textwidth]{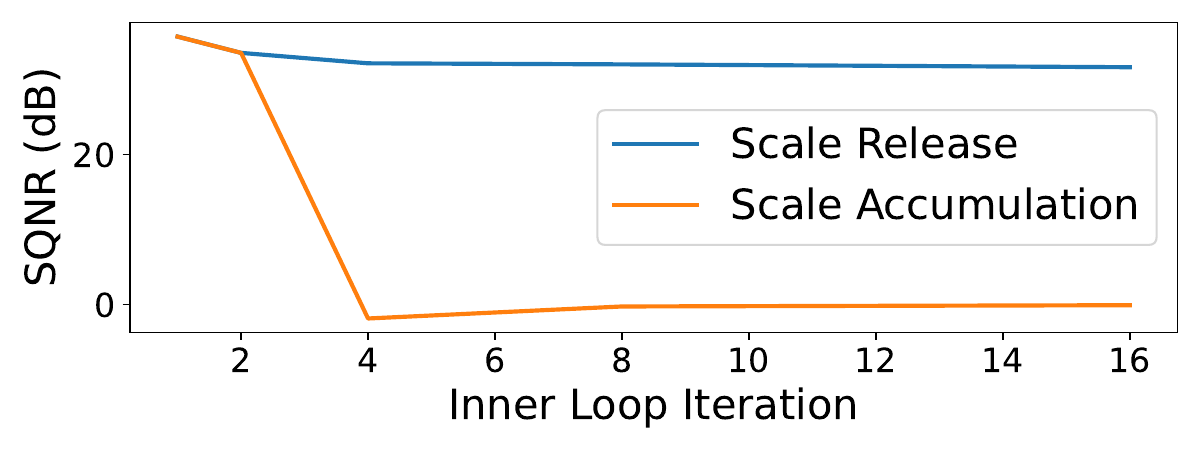}
\caption{Comparison of SQNR for scale accumulation vs. scale release.}
\label{fig:tile_sqnr}
\end{figure}

Figure~\ref{fig:tile_sqnr} compares the SQNR trends with respect to the number of inner-loop iterations. 
While both approaches exhibit decreasing SQNR as the iteration count increases, the scale accumulation approach shows a rapid SQNR collapse as accumulation progresses. 
In contrast, the scale release approach remains relatively stable even under repeated tile-wise accumulation. 
Therefore, we adopt the scale release approach as it is better suited for tile-based integer FlashAttention.

\subsection{C2: Implementation Details of GPU-Friendly Shiftmax}
\label{app:kernel_c2}

Shiftmax reduces the computational complexity of softmax by approximating its key operation, the exponential function, using integer arithmetic and bit-shifts instead of the floating-point $\exp$.
In particular, by converting the exponential into a base-2 form, the exponential computation can be replaced with a combination of multiplication and shift operations, enabling a hardware-friendly implementation.

In this work, we implement the base-2 exponential using only integer operations by decomposing the input $\hat{x}$ into a quotient ($q$) and a remainder ($r$), and generating the output $\hat{y}$ via shift operations.
Algorithm~2 summarizes the integer-based base-2 exponential procedure.
Here, $M$ is a constant quantized with a $2^N$ scaling factor for fixed-point multiplication, and on GPUs the computation consists only of integer multiplication (int $\times$ int) and shift operations.

A conventional method to compute $q$ is given as follows:
\begin{equation}
q = \left\lfloor \frac{\hat{x}}{-s_x^{(\text{inv})}} \right\rfloor
\label{eq:q_div}
\end{equation}

However, this method requires integer division, which is expensive on GPUs and may become a bottleneck in the overall runtime of ShiftExp2.
Therefore, instead of performing integer division directly, we approximate $q$ using integer multiplication and shift operations with a fixed-point constant $M$.
That is, we replace the division with a mul+shift form as follows.

\begin{equation}
M = \left\lfloor (-s_x)\cdot 2^N \right\rceil,\qquad
q = (\hat{x}\cdot M)\gg N
\label{eq:q_mulshift}
\end{equation}

Algorithm~2 describes the ShiftExp2 procedure, which approximates the base-2 exponential of the input $\hat{x}$ using only integer operations. 
In particular, to avoid expensive integer division on GPUs, the quotient $q$ is computed using integer multiplication and shift operations with a fixed-point constant $M$. 
The remainder $r$ is then derived, and the final output $\hat{y}$ is generated via shift-based operations.

\begin{algorithm}[H]
\caption{Integer-only Base-2 Exponential}
\label{algo:shiftexp2}
\begin{algorithmic}[1]
\STATE \textbf{Input:} $\hat{x}, s_x$: quantized input and scale
\STATE \textbf{Output:} $\hat{y}, s_y$: quantized output and scale
\STATE \textbf{function} ShiftExp2$(\hat{x}, s_x)$
\STATE \hspace{\algorithmicindent} $s_x^{(\text{inv})} \gets \left\lfloor 1 / s_x \right\rceil$
\STATE \hspace{\algorithmicindent} $M \gets \left \lfloor(-s_x) \cdot 2^N \right \rceil$
\STATE \hspace{\algorithmicindent} $q \gets (\hat{x} \cdot M)\gg N$
\STATE \hspace{\algorithmicindent} $r \gets \hat{x} + q \cdot s_x^{\text{(inv)}}$
\STATE \hspace{\algorithmicindent} $\hat{y} \gets ((r \gg 1) + s_x^{\text{(inv)}}) \gg q $
\STATE \hspace{\algorithmicindent} $s_y \gets s_x$
\STATE \hspace{\algorithmicindent} \textbf{return} $\hat{y}, \; s_y$

\STATE \textbf{end function}
\end{algorithmic}
\end{algorithm}

\begin{figure}[H]
\centering
\includegraphics[width=0.5\textwidth]{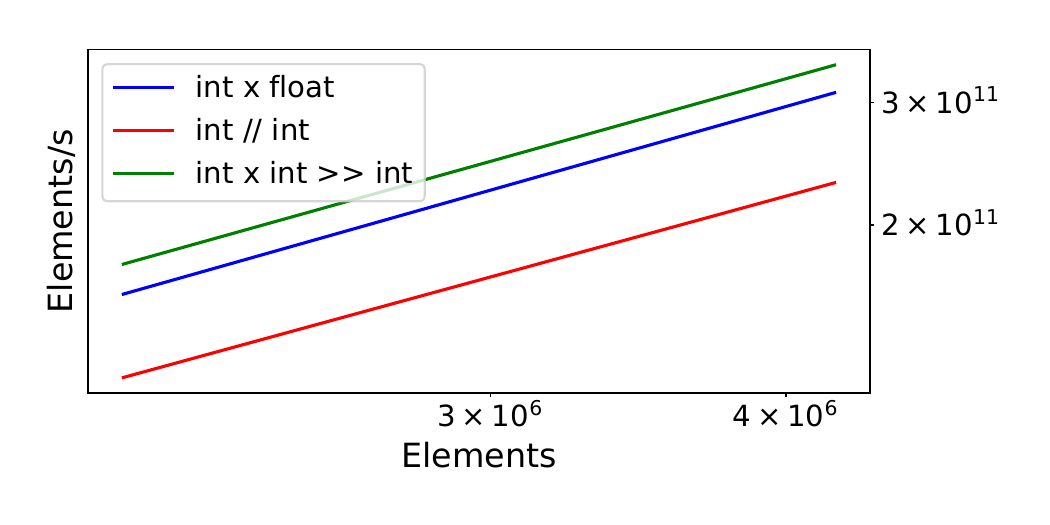}
\caption{Throughput comparison of three arithmetic operations.}
\label{fig:mul}
\end{figure}

Figure~\ref{fig:mul} compares the throughput of integer division and mul+shift operations on GPUs. 
Integer division exhibits significantly lower throughput than mul+shift, indicating that computing $q$ via integer division in ShiftExp2 can become a performance bottleneck. 
Therefore, in our ShiftExp2 implementation, we avoid direct integer division and instead compute $q$ using a mul+shift formulation based on fixed-point multiplication and shift operations.

\subsection{C3: Granularity Analysis for Integer Fused Attention}
\label{app:kernel_c3}

\begin{figure}[H]
\centering
\includegraphics[width=0.48\textwidth]{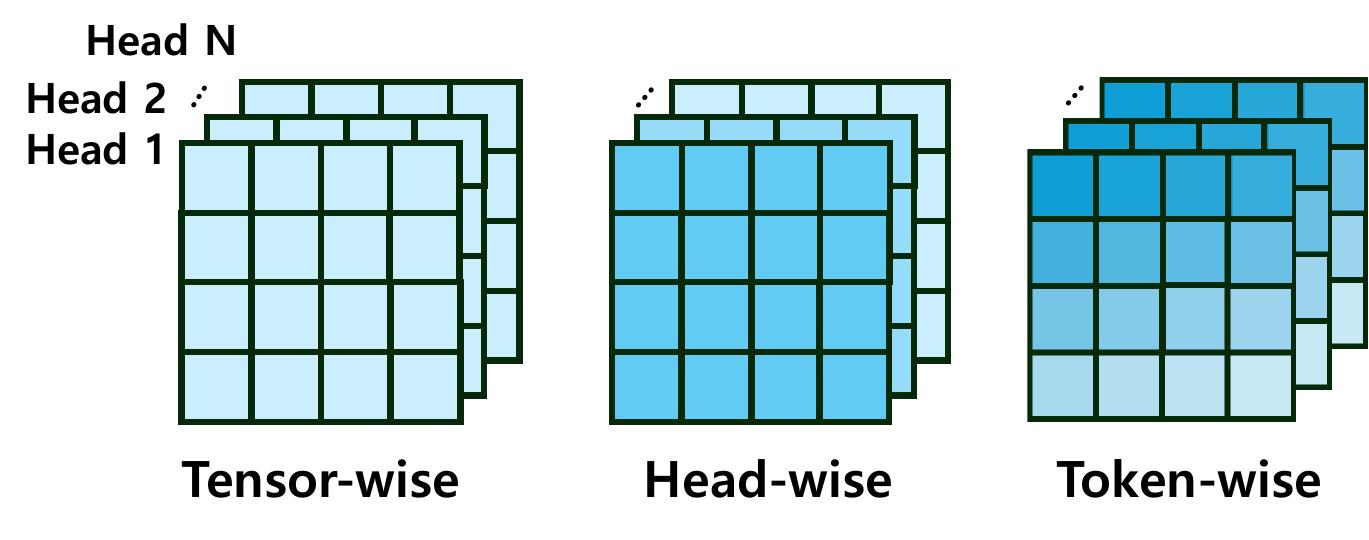}
\caption{Quantization granularity options for attention inputs.}
\label{fig:8}
\end{figure}

The input tensor for attention computation consists of the batch size, the number of heads, the sequence length, and the per-head dimension. 
For quantization granularity, we can choose among per-tensor, per-head, and per-token schemes, as illustrated in Figure~\ref{fig:8}.
Per-tensor quantization applies a single scale to the entire input tensor, offering the simplest implementation and the lowest scale storage cost. 
Per-head quantization assigns different scales to each head, capturing head-wise distribution differences and providing a balanced trade-off between accuracy and overhead. 
Per-token quantization applies a separate scale for each token, best reflecting distribution variations, but it introduces additional overhead for scale computation and application.

\begin{figure}[H]
\centering 
\includegraphics[width=0.48\textwidth]{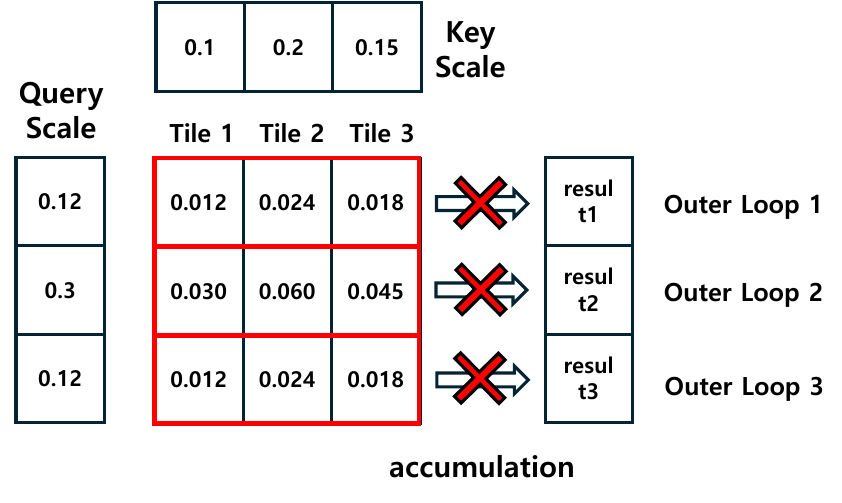}
\caption{Impossibility of integer accumulation with per-token granularity}
\label{fig:9}
\end{figure}

Figure~\ref{fig:9} illustrates that applying per-token granularity results in different scales across tiles of the QK transposed GEMM output, making integer-based accumulation difficult. In an integer fused attention kernel, the QK transposed GEMM output is computed tile by tile and repeatedly accumulated along the outer loop. 
For integer comparison and accumulation to remain semantically consistent, all accumulated tiles must share the same scale. If each tile has a different scale as shown in Figure~8, integer operations would effectively add or compare values represented in different units, which can introduce errors in the accumulated result. 
Therefore, when maintaining a simple integer-only execution path, the per-token scheme is limited in practice, and per-tensor or per-head granularity becomes a more feasible choice.

\end{document}

%% file: algorithm.tex

\begin{algorithm*}[t]
\caption{QFlash Algorithm}
\label{algo:1}
\begin{algorithmic}[1]
\STATE \textbf{Require:}
$\hat{\mathbf{Q}}, \hat{\mathbf{K}}, \hat{\mathbf{V}} \in \mathbb{I}_{\text{int8}}^{N \times d}$,
scale factors $s_Q, s_K, s_V \in \mathbb{R}$ and $s = s_Q \cdot s_K \cdot d^{-1/2} \cdot \log_2 e$,
block sizes $B_c, B_r$,
multiplier $M_r$,
shift bits $r$.
\STATE Divide $\hat{\mathbf{Q}}$ into $T_r = \lceil N/B_r \rceil$ blocks $\hat{\mathbf{Q}}_1, \dots, \hat{\mathbf{Q}}_{T_r}$ of size $B_r \times d$, and divide $\hat{\mathbf{K}}, \hat{\mathbf{V}}$ into $T_c = \lceil N/B_c \rceil$ blocks $\hat{\mathbf{K}}_1, \dots, \hat{\mathbf{K}}_{T_c}$ and $\hat{\mathbf{V}}_1, \dots, \hat{\mathbf{V}}_{T_c}$ of size $B_c \times d$ each.
\STATE Divide output $\hat{\mathbf{O}} \in \mathbb{I}_{\text{int}8}^{N \times d}$ into $T_r$ blocks $\hat{\mathbf{O}}_1, \dots, \hat{\mathbf{O}}_{T_r}$ of size $B_r \times d$ each.
\FOR{$i = 1$ \TO $T_r$}
    \STATE Load $\hat{\mathbf{Q}}_i$ from $\hat{\mathbf{Q}}[(i-1)B_r:iB_r, :]$ from off-chip to SRAM
    \STATE On chip, initialize $\hat{\mathbf{O}}_i = \mathbf{0} \in \mathbb{Z}^{B_r \times d}$, $\hat{l}_i^{(0)} = \mathbf{0} \in \mathbb{Z}^{B_r}$, $\hat{m}_i^{(0)} = -2^{21} \in \mathbb{Z}^{B_r}$
    \FOR{$j = 1$ \TO $T_c$}
        \STATE Load $\hat{\mathbf{K}}_j^\top$ from $\hat{\mathbf{K}}^\top[:, (j-1)B_c:jB_c]$ and $\hat{\mathbf{V}}_j$ from $\hat{\mathbf{V}}[(j-1)B_c:jB_c, :]$ from off-chip to SRAM
        \STATE On chip, compute $\hat{\mathbf{S}}_i^{(j)} \gets \hat{\mathbf{Q}}_i \hat{\mathbf{K}}_j^\top$ \hfill $\triangleright$ \circnum{1}
        \STATE On chip, compute $\hat{m}_i^{(j)} \gets \max(\hat{m}_i^{(j-1)}, \operatorname{rowmax}(\hat{\mathbf{S}}_i^{(j)}))$ \hfill $\triangleright$ \circnum{2}, \circnum{3}
        \STATE On chip, compute $\hat{\alpha}_i^{(j)} \gets \operatorname{ShiftExp2}(\hat{m}_i^{(j-1)} - \hat{m}_i^{(j)}, s)$ \hfill $\triangleright$ \circnum{4}
        \STATE On chip, compute $\hat{\mathbf{P}}_i^{(\text{tiled})} \gets \operatorname{ShiftExp2}(\hat{\mathbf{S}}_i^{(j)} - \hat{m}_i^{(j)}, s)$ \hfill $\triangleright$ \circnum{5}
        \STATE On chip, compute $\hat{\mathbf{P}}_i^{(j)} \gets \operatorname{Requantization}(\hat{\mathbf{P}}_i^{(\text{tiled})},\, M,\, b)$ \hfill $\triangleright$ \circnum{6}
        \STATE On chip, compute $\hat{l}_i^{(j)} \gets \operatorname{ScaleRelease}(\hat{l}_i^{(j-1)} ,\, \hat{\alpha}_i^{(j)},\, s) + \operatorname{rowsum}(\hat{\mathbf{P}}_i^{(j)})$ \hfill $\triangleright$ \circnum{7}, \circnum{9}
        \STATE On chip, compute $\hat{\mathbf{O}}_i^{(j)} \gets \operatorname{ScaleRelease}(\hat{\mathbf{O}}_i^{(j-1)} ,\, \hat{\alpha}_i^{(j)},\, s) + \hat{\mathbf{P}}_i^{(j)} \hat{\mathbf{V}}_j$ \hfill $\triangleright$ \circnum{8}, \circnum{10}
    \ENDFOR
    \STATE On chip, compute $\hat{\mathbf{O}}_i \gets \lfloor \hat{\mathbf{O}}_i^{T_c} / \hat{l}_i^{T_c} \rfloor$ \hfill $\triangleright$ \circnum{11}
    \STATE Write $\hat{\mathbf{O}}_i$ to off-chip as the $i$-th block of $\hat{\mathbf{O}}$
\ENDFOR
\STATE \textbf{Return} the output $\hat{\mathbf{O}} \in \mathbb{I}_{\text{int8}}^{N \times d}$ and the output scale factor $s_O = s_V$.

\end{algorithmic}
\end{algorithm*}


%% file: simple-base.bib
@inproceedings{zhao2025quark,
  title={QUARK: Quantization-Enabled Circuit Sharing for Transformer Acceleration by Exploiting Common Patterns in Nonlinear Operations},
  author={Zhao, Zhixiong and Li, Haomin and Liu, Fangxin and Lu, Yuncheng and Wang, Zongwu and Yang, Tao and Jiang, Li and Guan, Haibing},
  booktitle={2025 IEEE/ACM International Conference On Computer Aided Design (ICCAD)},
  pages={1--9},
  year={2025},
  organization={IEEE}
}

@article{jahadi2025fused,
  title={Fused Tensor Core: A Hardware--Software Co-Design for Efficient Execution of Attentions on GPUs},
  author={Jahadi, Reza and Munz, Phil and Atoofian, Ehsan},
  journal={IEEE Embedded Systems Letters},
  volume={17},
  number={5},
  pages={317--320},
  year={2025},
  publisher={IEEE}
}

@inproceedings{qin2025picachu,
  title={PICACHU: Plug-In CGRA Handling Upcoming Nonlinear Operations in LLMs},
  author={Qin, Jiajun and Xia, Tianhua and Tan, Cheng and Zhang, Jeff and Zhang, Sai Qian},
  booktitle={Proceedings of the 30th ACM International Conference on Architectural Support for Programming Languages and Operating Systems, Volume 2},
  pages={845--861},
  year={2025}
}

@techreport{leinhauser2021performance,
  title={Performance analysis of PIConGPU: particle-in-cell on GPUs using NVIDIA’s NSight systems and NSight compute},
  author={Leinhauser, Matthew and Young, Jeffrey and Bastrakov, Sergei and Widera, Ren{\'e} and Chatterjee, Ronnie and Chandrasekaran, Sunita},
  year={2021},
  institution={Oak Ridge National Laboratory (ORNL), Oak Ridge, TN (United States)}
}

@inproceedings{dally2023hardware,
  title={Hardware for deep learning},
  author={Dally, Bill},
  booktitle={2023 IEEE Hot Chips 35 Symposium (HCS)},
  pages={1--58},
  year={2023},
  organization={IEEE Computer Society}
}

@inproceedings{horowitz20141,
  title={1.1 computing's energy problem (and what we can do about it)},
  author={Horowitz, Mark},
  booktitle={2014 IEEE international solid-state circuits conference digest of technical papers (ISSCC)},
  pages={10--14},
  year={2014},
  organization={IEEE}
}

@inproceedings{yu2023know,
  title={Know your enemy to save cloud energy: Energy-performance characterization of machine learning serving},
  author={Yu, Junyeol and Kim, Jongseok and Seo, Euiseong},
  booktitle={2023 IEEE International Symposium on High-Performance Computer Architecture (HPCA)},
  pages={842--854},
  year={2023},
  organization={IEEE}
}

@article{bridges2016understanding,
  title={Understanding GPU power: A survey of profiling, modeling, and simulation methods},
  author={Bridges, Robert A and Imam, Neena and Mintz, Tiffany M},
  journal={ACM Computing Surveys (CSUR)},
  volume={49},
  number={3},
  pages={1--27},
  year={2016},
  publisher={ACM New York, NY, USA}
}

@article{you2024shiftaddllm,
  title={Shiftaddllm: Accelerating pretrained llms via post-training multiplication-less reparameterization},
  author={You, Haoran and Guo, Yipin and Fu, Yichao and Zhou, Wei and Shi, Huihong and Zhang, Xiaofan and Kundu, Souvik and Yazdanbakhsh, Amir and Lin, Yingyan Celine},
  journal={Advances in Neural Information Processing Systems},
  volume={37},
  pages={24822--24848},
  year={2024}
}

@inproceedings{zhong2024erq,
  title={Erq: Error reduction for post-training quantization of vision transformers},
  author={Zhong, Yunshan and Hu, Jiawei and Huang, You and Zhang, Yuxin and Ji, Rongrong},
  booktitle={Forty-first International Conference on Machine Learning},
  year={2024}
}

@inproceedings{ptq4vit,
  title={Ptq4vit: Post-training quantization for vision transformers with twin uniform quantization},
  author={Yuan, Zhihang and Xue, Chenhao and Chen, Yiqi and Wu, Qiang and Sun, Guangyu},
  booktitle={European conference on computer vision},
  pages={191--207},
  year={2022},
  organization={Springer}
}

@inproceedings{mixed_v1,
  title={Mixed Non-linear Quantization for Vision Transformers},
  author={Kim, Gihwan and Lee, Jemin and Park, Sihyeong and Kwon, Yongin and Kim, Hyungshin},
  booktitle={European Conference on Computer Vision},
  pages={97--112},
  year={2024},
  organization={Springer}
}

@inproceedings{repq-vit,
  title={Repq-vit: Scale reparameterization for post-training quantization of vision transformers},
  author={Li, Zhikai and Xiao, Junrui and Yang, Lianwei and Gu, Qingyi},
  booktitle={Proceedings of the IEEE/CVF International Conference on Computer Vision},
  pages={17227--17236},
  year={2023}
}

@inproceedings{clamp-vit,
  title={Clamp-vit: Contrastive data-free learning for adaptive post-training quantization of vits},
  author={Ramachandran, Akshat and Kundu, Souvik and Krishna, Tushar},
  booktitle={European Conference on Computer Vision},
  pages={307--325},
  year={2024},
  organization={Springer}
}

@inproceedings{kim2021bert,
  title={I-bert: Integer-only bert quantization},
  author={Kim, Sehoon and Gholami, Amir and Yao, Zhewei and Mahoney, Michael W and Keutzer, Kurt},
  booktitle={International conference on machine learning},
  pages={5506--5518},
  year={2021},
  organization={PMLR}
}

@article{fq-vit,
  title={Fq-vit: Post-training quantization for fully quantized vision transformer},
  author={Lin, Yang and Zhang, Tianyu and Sun, Peiqin and Li, Zheng and Zhou, Shuchang},
  journal={arXiv preprint arXiv:2111.13824},
  year={2021}
}

@article{illm,
  title={I-llm: Efficient integer-only inference for fully-quantized low-bit large language models},
  author={Hu, Xing and Cheng, Yuan and Yang, Dawei and Yuan, Zhihang and Yu, Jiangyong and Xu, Chen and Zhou, Sifan},
  journal={arXiv preprint arXiv:2405.17849},
  year={2024}
}

@article{arafa2019instructions,
  title={Instructions’ latencies characterization for nvidia gpgpus},
  author={Arafa, Yehia and Badawy, AH and Chennupati, Gopinath and Santhi, Nandakishore and Eidenbenz, Stephan},
  journal={arXiv preprint arXiv:1905.08778},
  year={2019}
}

@article{dao2022flashattention,
  title={Flashattention: Fast and memory-efficient exact attention with io-awareness},
  author={Dao, Tri and Fu, Dan and Ermon, Stefano and Rudra, Atri and R{\'e}, Christopher},
  journal={Advances in neural information processing systems},
  volume={35},
  pages={16344--16359},
  year={2022}
}

@article{dao2023flashattention,
  title={Flashattention-2: Faster attention with better parallelism and work partitioning},
  author={Dao, Tri},
  journal={arXiv preprint arXiv:2307.08691},
  year={2023}
}

@article{shah2024flashattention,
  title={Flashattention-3: Fast and accurate attention with asynchrony and low-precision},
  author={Shah, Jay and Bikshandi, Ganesh and Zhang, Ying and Thakkar, Vijay and Ramani, Pradeep and Dao, Tri},
  journal={Advances in Neural Information Processing Systems},
  volume={37},
  pages={68658--68685},
  year={2024}
}

@inproceedings{li2023vit,
  title={I-vit: Integer-only quantization for efficient vision transformer inference},
  author={Li, Zhikai and Gu, Qingyi},
  booktitle={Proceedings of the IEEE/CVF International Conference on Computer Vision},
  pages={17065--17075},
  year={2023}
}

@inproceedings{kluska2024qattn,
  title={Qattn: Efficient gpu kernels for mixed-precision vision transformers},
  author={Kluska, Piotr and Castell{\'o}, Adri{\'a}n and Scheidegger, Florian and Malossi, A Cristiano I and Quintana-Ort{\'\i}, Enrique S},
  booktitle={Proceedings of the IEEE/CVF Conference on Computer Vision and Pattern Recognition},
  pages={3648--3657},
  year={2024}
}

@article{chen2024int,
  title={Int-flashattention: Enabling flash attention for int8 quantization},
  author={Chen, Shimao and Liu, Zirui and Wu, Zhiying and Zheng, Ce and Cong, Peizhuang and Jiang, Zihan and Wu, Yuhan and Su, Lei and Yang, Tong},
  journal={arXiv preprint arXiv:2409.16997},
  year={2024}
}

@article{milakov2018online,
  title={Online normalizer calculation for softmax},
  author={Milakov, Maxim and Gimelshein, Natalia},
  journal={arXiv preprint arXiv:1805.02867},
  year={2018}
}
